\documentclass[format=acmtog, review=false, anonymous=false, timestamp=true, nonacm=true]{acmart}
\pdfoutput=1

\usepackage[T1]{fontenc}
\usepackage[utf8]{inputenc}
\usepackage[english]{babel}

\usepackage{bm}

\usepackage{standalone}

\usepackage{amsmath}
\usepackage{amsthm}
\theoremstyle{definition}
\theoremstyle{plain}
\theoremstyle{remark}

\DeclareMathAlphabet{\mathcalligra}{T1}{calligra}{m}{n}

\usepackage{siunitx}

\usepackage{tabu}
\usepackage{booktabs}

\usepackage{xcolor}
\usepackage{graphicx}
\usepackage{pgfbaseimage}

\usepackage{glossaries}

\usepackage{cleveref}

\usepackage[disable]{todonotes}

\usepackage{subcaption}

\graphicspath{{img/}, {figures/}, {~/share/img/}}

\newcommand{\ie}{\mbox{i.\,e.,}}

\newcommand{\eg}{\mbox{e.\,g.,}}
\newcommand{\Eg}{\mbox{E.\,g.,}}

\newcommand{\R}{\ensuremath{\mathbb{R}}} 

\newcommand{\mA}{\ensuremath{\mathcal{A}}}  
\newcommand{\mM}{\ensuremath{\mathcal{M}}}  

\newcommand{\mathvf}[1]{\ensuremath{\bm{#1}}} 

\newcommand{\todolaf}[1]{\todo[color=orange!40]{LaF: #1}}
\newcommand{\todoeve}[1]{\todo[color=green!40]{EVe: #1}}

\makeglossaries

\newacronym{AL}{AL}{Adversarial Learning}
\newacronym{ANN}{ANN}{Artificial Neural Network}
\newacronym{ARL}{ARL}{Adversarial Resilience Learning}
\newacronym{GAN}{GAN}{Generative Adversarial Network}
\newacronym{GPGPU}{GPGPU}{General-Purpose Graphics Processing Unit}
\newacronym{DL}{DL}{Deep Learning}
\newacronym{GRU}{GRU}{Gated Recurrent Unit}
\newacronym{ICT}{ICT}{Information and Communication Technology}
\newacronym{LSTM}{LSTM}{Long-Short Term Memory}

\newacronym{ML}{ML}{Machine Learning}
\newacronym{PoC}{PoC}{Proof-of-Concept}
\newacronym{RL}{RL}{Reinforcement Learning}
\newacronym{RNN}{RNN}{Recurrent Neural Network}
\newacronym{SIMD}{SIMD}{Single Instruction, Multiple Data}

\newglossaryentry{resilience}{
  name=resilience,
  description={The ability of a system to withstand unforseen, rare and
  potentially catastrophic events and to self-heal or improve in
  reaction to these events. Ideally resilience is increasing monotonously.}
}

\newacronym{AEG}{AEG}{Attack Execution Graph}

\setcopyright{none} 
\acmConference[Anonymous Submission to HSCC 2019]{HSCC 2019}{Due 03 July 2018}{Montreal, Canada}
\acmYear{2018}

\begin{document}

\title{Adversarial Resilience Learning 
  --- Towards Systemic Vulnerability Analysis for Large and Complex Systems}
\author{Lars Fischer}\email{lars.fischer@offis.de}
\author{Jan-Menno Memmen}
\author{Eric MSP Veith}\email{eric.veith@offis.de}
\author{Martin Tr\"oschel}\email{martin.troeschel@offis.de}

\begin{abstract}

  This paper introduces \gls{ARL}, a concept to model, train, and analyze
  artificial neural networks as representations of competitive agents in
  highly complex systems. In our examples, the agents normally take the roles
  of attackers or defenders that aim at worsening or improving---or keeping,
  respectively---defined performance indicators of the system. Our concept
  provides adaptive, repeatable, actor-based testing with a chance of
  detecting previously unknown attack vectors. We provide the constitutive
  nomenclature of \gls{ARL} and, based on it, the description of experimental
  setups and results of a preliminary implementation of \gls{ARL} in simulated
  power systems. 

\end{abstract}


\maketitle
\glsresetall

\section{Introduction}
\label{sec:introduction}

\todoeve{Better start: this is very abrupt. Add some intro text.}

Current newspapers are full of horrific tales of ``cyber-attackers''
threatening our energy systems. And, if not for the notorious ``evil
state''-actor, it is the ongoing digitization necessary to enable increasing
renewable and volatile energy generation that threatens our energy supply and
thus the stability of our society. And while the main approach seems to be to
patch-up the detected vulnerabilities of protocols, software and controller
devices, our approach is to research and develop the means to systematically
design and test systems that are structurally resilient against failures and
attackers alike.


Security in cyber-systems mostly should be concerned with establishing
asymetric control in favour of the operator of a system. In order to achieve
this on a structural level at design time, reproducible benchmark tests are
required. This is notoriously difficult for intelligent adversaries whose
primary ability are adaption and creativity. Thus, testing methods nowadays
are either reproducible and insufficiently modelling attacker — or they
involve unreproducible human elements. Reinforcement learning may be useful to
provide at least some adaptability of reproducible attacker models.

This work takes its motivation and first practical implementation from the
power system domain, but the work can directly be applied to all highly
complex, critical systems. Systems that may benefit from \gls{ARL} are too
complex to be sufficiently described using analytic methods, \ie{} because the
number of potential states is to large and the behaviour is too complex with
too many non-trivial interdependencies.

This work introduces \gls{ARL}, which provides a method to analyse complex
interdependent systems with respect to adversarial actors. The foremost
motivation is to provide a method for deterministic analysis method for
complex and large systems including some degree of adaptivity of the simulated
attackers. 

The main contribution of this paper is the introduction of a novel structure
for training and application of \glspl{ANN} that generalizes the approach of
adversarial learning. By setting up ANN-based agents in a competitive
situation, the learning-complexity is comprised not only of a highly complex
system, but also of competing \glspl{ANN} whose changing state, manifested by
modified behaviour of the system under consideration, has to be included in
the trained model. We assume that this provides a very interesting new problem
class for \gls{RL}, as it introduces a cyclic learning competition. 

The paper is structured as follows. First, a brief introduction into related
techniques in machine learning and related work for complex system analysis
is given. The paper then defines the concept of Adversarial Resilience
Learning in \Cref{sec:advers-resil-learn}, and introduces its
application to adversary testing in power system control in
\Cref{sec:appl-power-syst}. The paper is completed by a presentation of
lessons' learned and results from an early proof-of-concept demonstrator in
\Cref{sec:demonstrator}. Itconcludes with a discussion and an outlook
in \Cref{sec:disc-future-work}

\section{Related Work}
\label{sec:related-work}

This work aims at exploring the feasibility of improving resilience of
complex systems using machine learning to train adaptive agents. The
term \emph{\gls{resilience}} is lacking a coherent and precise
definition across fields.  Generally it denotes the ability of a
system to withstand unforseen, rare and potentially catastrophic
events, recover from the damage and adapt by improving itself in
reaction to these events. Ideally, resilience is increasing
monotonously throughout system improvement. A useful simplification is
observation of the changing behaviour of system performance as an
artefact resulting from resilience processes.  Different formalisation
of resilience processes exist, but most distinguish subprocesses for
planning, absorbtion of damage, recovery (or self-healing) and
improvement (or adaption).~\cite{arghandeh2016cp_resilience}

\begin{figure}
  \centering
  \includegraphics[width=\linewidth]{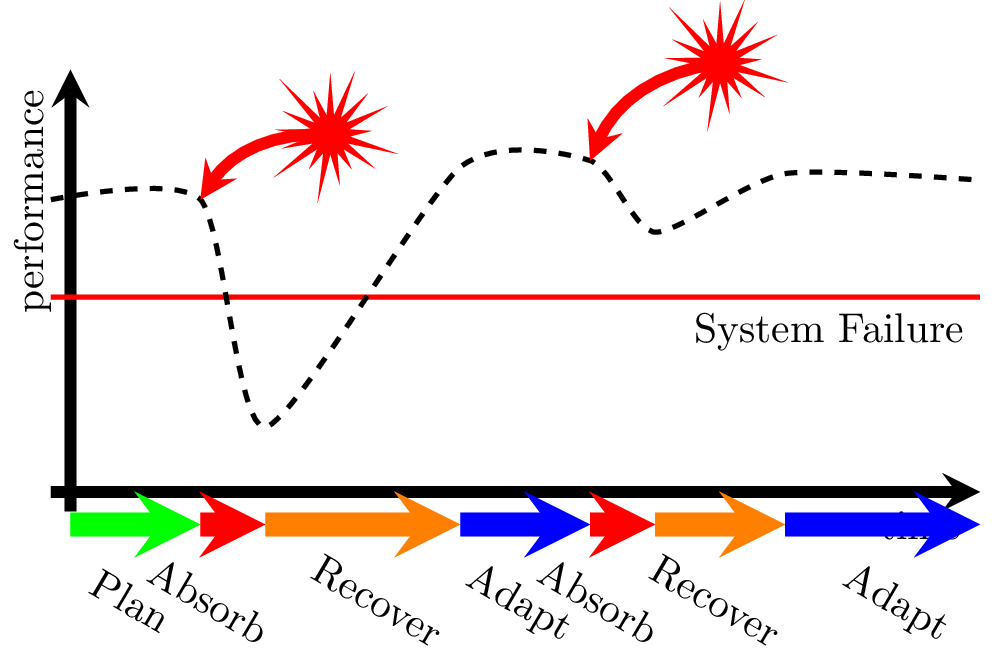}
  \caption{Resilience Process for system performance}
  \label{fig:resilience}
\end{figure}

See \Cref{fig:resilience} for an expression of a hypothetical
system's performance suffering twice from damaging
events. \Gls{resilience} is modelled as a sequential process: Plan,
Absorb, Recover and Adapt.~\cite{Linkov2019} As
consequence of the first event the performance of the system is pushed
below a \emph{failure threshold}, \ie{} the system fails to provide
its service. Improvement of the system is then achieved after recovery
as the system is able to keep the performance above the failure
threshold during the second event. 

\subsection{Analysis and Stochastic Modelling}
\label{sec:formal-models}


The main distinction of our approach as compared to game theoretic
modelling and stochastic analysis is the use of co-simulation and
heuristic approaches instead of formal abstraction of complete
systems. The underlying assumption is, that a system-of-systems is too
complex, and malicious adversaries are too unpredictable to be
sufficiently analysed.

But similarly to Attacker-Defender Models, \eg{} described in
\cite{brown2006defending_critical_infrastructures} that at analysing an
equilibrium between attackers and defenders in dynamic systems, our work aims
at heuristically approach an estimate of the asymmetry of attacker and
defender in these systems. The approach of \gls{ARL} structurally similar to
the concept of \emph{Stackleberg Competition} and related applications of
stochastic analysis, \eg{} pursuit-evasion in differential
games~\cite{isaacs1954differential}. These approaches seem to only be
applicable to scenarios that can be restricted to few degrees of freedom. More
realistic behaviours of opportunistically acting threat agents within complex
system-of-systems leads to an explosion of states in analytic approaches.  

Recent surveys seem to support this view. Referenced approaches on power
systems in \cite{Do2017gametheoryforcybersec} provide no details on the used
game-theoretic model and use ambiguous terminology of the researched threat
scenarios. Approaches in \gls{ML} to tackle complex problems, on the other
hand, have, not only recently, been very successful in providing novel,
practical solutions.

\subsection{Machine Learning}
\label{sec:machine-learning}


\glspl{ANN} are universal function approximators, meaning that they
can be used as a statistical model of any Borel-measurable function
\(\mathbb{R}^n \mapsto \mathbb{R}^m \) with desired non-zero error
\citep{cybenko1989approximation,hornik1989multilayer,goodfellow2016deep}.
Already the standard \gls{RNN} (\eg{} \citet{elman1990finding}) has
the capacity to approximate any non-linear dynamic system
\citep{seidl1991dynamicsystem}, and Siegelmann and Sonntag have shown
that \glspl{RNN} are turing-complete
\citep{siegelmann1995computational}.

In practice, a typical problem for which \glspl{RNN}, especially structures
containing \gls{LSTM} cells \citep{hochreiter1997long} or \glspl{GRU}
\citep{cho2014learning,chung2014empirical} are used, is time series
prediction. Even greater memory capacity is achieved by \emph{neural turing
machines}, introduced by Graves et.~al. \citep{graves2014neural}, which also
counter the learning problem that is inherent to the turing-completeness of
\glspl{RNN}: in theory, a \glspl{RNN} have the capacity to simulate arbitrary
procedures, given the proper set of parameters; in practice, this training
task has proven to be complicated.  Neural Turing Machines counter the
complexity with a vastly increased addressable memory space and have shown to
be able to simulate simple, but complete algorithms like sorting
[\emph{ibd.}].

Predicting a time series with an \gls{RNN} constitutes the instantiation of a
(non-linear) dynamic system
\citep{tong1990non,basharat2009chaoticmodeling,yu2007parameter}, \ie{} the
prediction is the result of the system's behavior, which is, in turn, modeled
and approximated by the \gls{RNN}. Cessac has examined \glspl{ANN} from the
perspective of dynamical systems theory, characterizing also the collective
dynamics of neural network models \citep{cessac2010view}.

Different ways exist to train \glspl{ANN} and \glspl{RNN}. When using
supervised training methods, the training set consists of both, vectors of
input and known output the \gls{ANN} is expected to exhibit. Two different
classes of training algorithms are popular for this type of training, with
gradient-decent-based algorithms of the Backpropagation-of-Error family
leading by far
\cite{rumelhart1986learning,kingma2014adam,dozat2016incorporating,ruder2016overview,cuellar2007application},
followed by evolutionary algorithms such as CMA-ES
\citep{hansen2006cma,ros2008simple,hansen2003reducing} or REvol
\citep{ruppert2014evolutionary,veith2017universal}. However, all optimization
methods adapt the \gls{ANN} to minimize a cost function and not directly to
create a model of a problem; this happens only indirectly. As a result,
\glspl{ANN} can still be ``foiled,'' \ie{} made to output widely wrong results
in the face of only minor modifications to the input. This effect and how to
counter it is the subject of adversarial learning research.

With unsupervised learning, the \gls{ANN} tries to detect patterns in the
input data that diverge from the background noise. Unsupervised learning does
not use the notion of expected output \citep{ghahramani2004unsupervised}. A
modern application of unsupervised learning has emerged in the concept of
\glspl{GAN}. Here, one network, called the generator network, creates
solution candidates---\ie maps a vector of latent variables to the solution
space---, which are then evaluated by a second network, the
discriminator \citep{goodfellow2014generative}. Ideally, the result of the
training process are results virtually indistinguishable from the actual
solution space, which is the reason \glspl{GAN} are sometimes called ``Turing
learning.'' The research focus of \gls{ARL} is not the generation of realistic
solution candidates; this is only a potential extension of the attackers and
defenders themselves. \gls{ARL}, however, describes the general concept of two
agents influencing a common model but with different sensors (inputs) and
actuators (output) and without knowing of each others presence or actions.

\Gls{RL} \citep{baird1994advanceupdate} describes a third class of learning
algorithms. It extends the process of adapting the weight matrix of any
\gls{ANN} by including the notion of an environment into the training process.
In \gls{RL}, an agent interacts with its environment that provides it with
feedback, which can be positive or negative. The agent's goal is to maximize
its reward. The agent internally approximates a reward function in order to
achieve a high reward through actions for every state of its environment.
Initially, \gls{RL} was not tied to \glspl{ANN} \citep{minsky1954neural}.
However, as the environment becomes more complex, so does the agent's reward
function; \glspl{ANN} are very well suited for function approximation or, if
the reward is based on a complex state of a world, \ie{} a dynamic system,
\glspl{RNN} are suitable.

For \gls{ARL}, we assume a common model that is used by two distinct agents:
while one probes the model for weaknesses in order to find attack vectors, the
other monitors the system and, unbeknowing of the presence of the attacker or
its actions, works at keeping the system in its nominal state. Through this
structure, the notion of \gls{ARL} assumes that the model---\ie{} each agent's
environment---is not completely known to the respective agent. Therefore, the
usage of \gls{RL} readily suggests itself.

The abstract notion of a model can see multiple instantiations; one such
instantiation of \gls{ARL} would be using a power grid as the model considered
by both agents. Ernst et~al. employ \gls{RL} for stability
control in power grids \cite{ernst2004stabilitycontrolreinforcementlearning}.
In their paper, they design a dynamic brake controller to damp large
oscillations; however, since the reward function is easily well-defined, there
is no need for using an \gls{ANN} for function approximation.

\begin{figure}
  \centering
  \includegraphics[width=\linewidth]{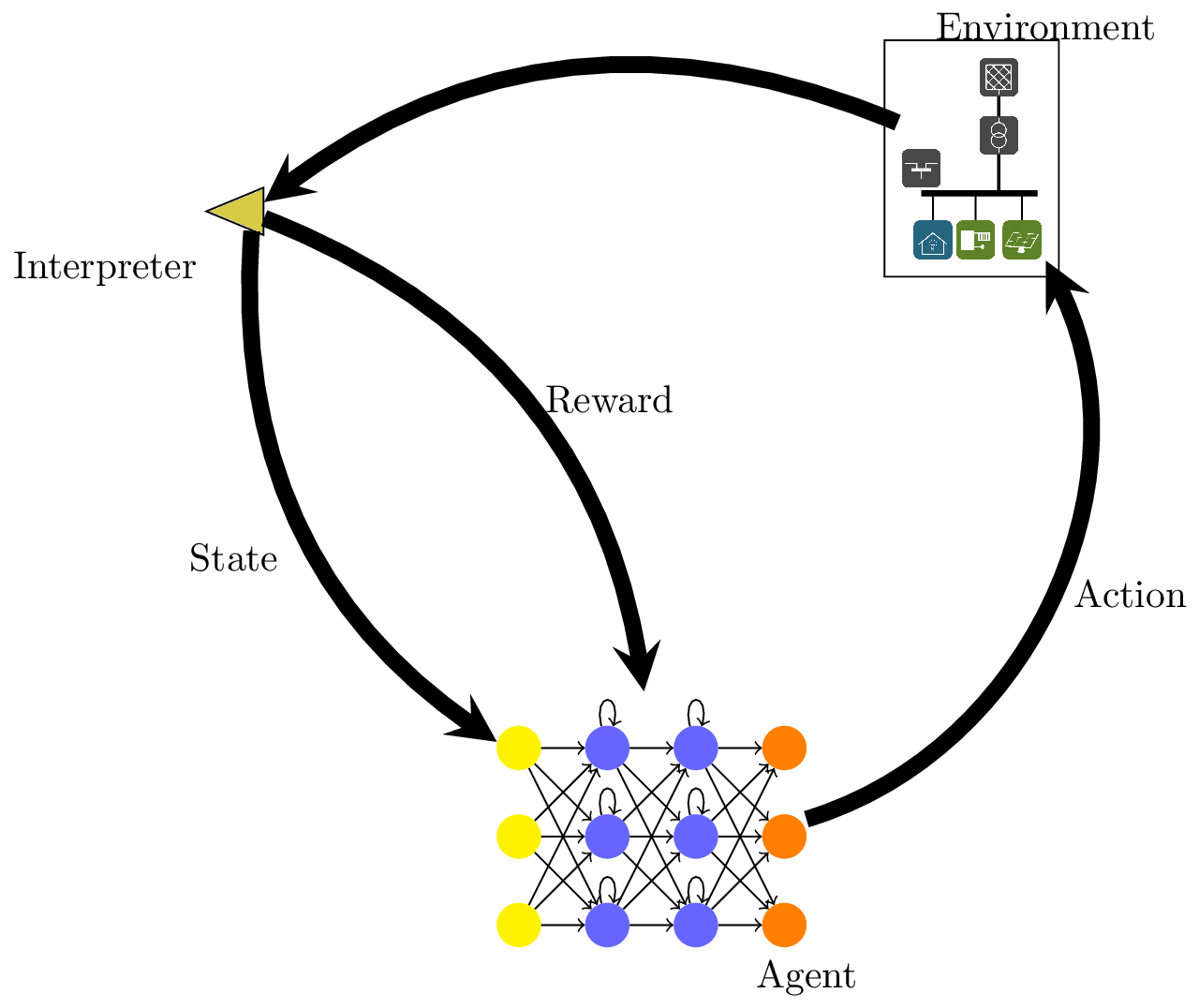}
  \caption{Reinforcement Learning}
  \label{fig:reinforcementlearning}
\end{figure}

\Cref{fig:reinforcementlearning} shows the general schema of \gls{RL}, adapted
from literature for the power grid as the environment.

The concept we propose in this paper is related to \gls{AL} insofar, as both
concepts use two distinct \glspl{ANN} with conflicting objectives
\citep{goodfellow2015harnessing_adversarial_examples}.

\Gls{AL} is the field of exploiting vulnerabilities in learning algorithms in
order to affect the behaviour of the resulting (learned) system. The prevalent
technique is specially crafting inputs that lead to unintended patterns being
recognised. A secondary objective is to use inputs that are not recognised as
disturbing patterns by humans. \Gls{AL} thus are a concept to implement a
Stackelberg Competition in neural networks.

A very recent example is given by Evtimov et~al. \cite{evtimov2017robust},
where the authors showed how they were able to confuse known deep learning
image recognition algorithms by attaching markings to physical objects.

\Gls{AL} neural network setups (for non-physical interventions) are
characterised by two output layers in sequence. The first output layer
functions as the adversary, generating outputs from a given input
layer, \ie{}
the ``real world inputs'' The second output layer represents the function of
the original classifier that has to learn to correctly classify the original
inputs despite the adversary's efforts to scramble the input. 

\section{Adversarial Resilience Learning}
\label{sec:advers-resil-learn}

\Gls{ARL} is distinguished from \gls{AL} by the recurrent structure in which
adversary and defender are interacting. While \gls{GAN}
directly connect a generating adversary with a detecting defender, \gls{ARL}
adversary and defender interact only through the system they are using for
input and output. In this interaction adversaries are identified as
agents inserting disturbances into the system, while defenders provide
resilience control.
 
\begin{definition}[Adversarial Resilience Learning (informal)]

  \gls{ARL} is an experimental structure comprised of two disjoint groups of
  agents and a system or simulated system. The agents are distinguished as
  attacker and defender by adhering to conflicting optimisation objectives.
  Both groups of agents receive their input from a, potentially overlapping,
  set of measurements from the system. They influence the system through two
  disjunct sets of outputs connected to controls in the simulated system.

\end{definition}

\subsection{Fundamental Notation and Model}
\label{sec:notation}

The basic abstract scenario using \gls{ARL} consists of two competing agents
and a system model. Each of the three elements resembles a state transition.
In order to establish a sound formal base, a definition of notation and
processes of \gls{ARL} is provided here. A summary of notations used is given
in \cref{tab:notation}.

\gls{ARL} consists of a set of agents, where each agent has a model, denoted
by \(\mA\), and a model of a system, \(\mM\). The agent model \(\mA\) serves
as a ``blue-print'' for the actual behavior of a running system; similarly,
\(\mM\) denotes a static model of a world. An index identifies a particular
agent model, \eg{} \(\mA_A\) denotes the category of attacker models,
\(\mA_\Omega\) serves to denote the category of defender models. At run-time,
the models are instantiated. We denote instances of a model with lower-case
letters \(a\), where the index denotes a particular state of the model, such
as \(a_t\) with $t$ commonly referring a point in simulation time. In the same
vein, \(m\) denotes an instance of a world model.


Each agent tries to maximize its rewards by approximating the agent-specific
performance function, 

\begin{equation}
  \label{eq:agent-performance}
  p_a(m_t)~.
\end{equation}
 
For an agent, the performance function \(p_a(\cdot)\) is equal to its reward
function in \gls{RL} terminology.  However, the notion of the
\emph{performance} function lets us decouple agent behavior from the
desired/intended or undesired performance of the world, denoted by

\begin{equation}
  \label{eq:model-performance}
  p(m_t)~,
\end{equation}

\noindent as the difference between the world's current performance to its
nominal performance, \(p^*\).

Agents are categorized through their performance function, an agent
model is identified as attacker model $\mA_A$ if his reward function
$p_a$ behaves inverse to the systems performance. The opposite is true
for agents from $\mA_\Omega$. That is, we can define:

\begin{definition}[Attacker and Defender Classes]
  \label{def:attacker_defender_classes}
  For all times $t$ and model instances $m\in\mM$, the
  following provides a classification rule for attackers and defenders:
  \begin{equation}
    \label{eq:attacker_reward}
    \begin{array}{rcl}
      a\in\mA_A & \Rightarrow p_a(m_t)& \nsim p(m_t),\\
      a\in\mA_\Omega & \Rightarrow p_a(m_t)& \sim p(m_t)~.
    \end{array}
  \end{equation}
\end{definition}
  
The performance of an agent is tightly coupled to an agent's view of its
environment, \ie{} the world. Each \(p_a(\cdot)\) can only be defined in terms
of the agent's sensory inputs\todoeve{it is possible to have performance
untainted by sensor dilution $\psi$}, \ie{} the part the world it can observe.
The portion of the state of a system instance an agent \(a\) can observe is
denoted by

\begin{equation}
  \label{eq:agent-input}
  x_{a,t} = \psi_a(m_t)~.
\end{equation}

Given the sensory inputs of an agent \(a\) at \(t\) are given as \(x_{a,t}\),
the agent can act by approximating its reward function \(p_a(\cdot)\). This
approximation is the agent's activation of its internal dynamic system
approximator \(\mathrm{act}(\cdot)\); implemented through, \eg{} an \gls{RNN},
such that

\begin{equation}
  \label{eq:act}
  \mathrm{act}_a: (a_t, x_{a,t}) \mapsto (a_{t+1}, y_{a,t+1})~,
\end{equation}


\noindent where we assume that an agent is always able to observe its
environment through \(\psi_a(m_t)\), \ie{} \(|x_{a,t}| > 0\), whereas it can
choose not to act, \ie{} \(| y_{a,t+1} | \ge 0 \).

An agent then acts on the system model through its actuators' actions,
\(y_{a,t}\), which are defined in its respective agent model, where each agent
has a specified set of actions of an actuator available at any given time.
Each agent defines an action policy for controlling its actuators. In the
simple case, the actions of an actuator are mapped from labels \([ y_1, y_2,
\dotsc, y_m ]^\top\) of, \eg{} an internal \gls{RNN}: Each evaluation step of
the performance function maps the sensor inputs \(x_{a,t}\) onto likelihood
values \(y_{a,t}\) for all labels of all actuators. The common interpretation
is that from each group of labels, the action mapped from the highest-valued
label is chosen as the one exerted onto the system by the agent. However
generally, an action policy takes on a form that is suitable for the whole
action search space, such as a policy network steering a monte carlo tree
search as has been shown in \cite{silver2017alphaGoZero}.  Thus, an agent is
acting through the evaluation and application of its system approximator.
This happens for each agent \(1, 2, \dotsc, n\). In brief, the systems
behavior is heavily influenced by the set of all actuators that can be
controlled by the respective agents. Thus, an agent does not simple perceive a
model (or a part thereof), but the state of the model as the result of all
agents acting upon it. Thus, an agent does not simply create an internal
representation of a dynamic system, but of a dynamical system-of-systems.

Finally, the simulator evaluates the actions of all agents applied to the
world model at \(t\), \(m_t\). This is represented by the evaluation function,

\begin{equation}
  \label{eq:eval}
  \mathrm{eval}: (\mathvf{y}_t, m_t) \mapsto m_{t+1}~,
\end{equation}

\noindent where the aggregated inputs and outputs over all agents are
represented as vectors
\begin{equation}
  \label{eq:io-vectors}
\mathvf{x}_t=
              \begin{bmatrix}
                x_{a_1,t}\\
                \vdots\\
                x_{a_n,t}\\
              \end{bmatrix},\quad
  \mathvf{y}_t =
                 \begin{bmatrix}
                   y_{a_1,t}\\
                   \vdots\\
                   y_{a_n,t}\\
                 \end{bmatrix}~.
\end{equation}

If the activation vectors of the participating agents consider a disjoint set
of controllers, \ie{} the actions application is commutative, the transition
of the world state from \(m_t\) to \(m_{t+1}\) is the result of an aggregation
of all agents' actions \(\mathvf{y}_t\). Non-commutative application of
actions is out of scope of this work.

\begin{table}
  \centering%
  \begin{tabu} to \linewidth {lX}
    \toprule
    \rowfont\bfseries Symbol & Description\\
    \midrule
    \(m\) of \(\mM\) & An instance of a system model\\
    \(a\) of \(\mA\) & An instance of an agent model\\
    \( \mA_A, \mA_\Omega \) & Attacker model, defender model (~\Cref{def:attacker_defender_classes})\\
    \( p(\cdot) \in \R_+ \) & Performance function\\
    \( p^*, p^f \) & Reference performance of normal operation,
      of failure threshold\\
    \( p(m_t) \) & Overall performance of a system instance \(m\)
      at time \(t\) (\cref{eq:model-performance})\\
    \( p_a(m_t) \) & Performance with respect to the objectives of agent
      instance \(a\) at \(t\) given the system instance \(m\) (\cref{eq:agent-performance})\\
    \( \psi_a\) & Observation function mapping a system model $m_t$
    onto the inputs available to agent $a$, \cref{eq:agent-input}\\
    \( x_{a,t}\) & Inputs to agent $a$ at time $t$, \cref{eq:agent-input}\\
    \( y_{a,t} \) & Actions \(y\)
    of \(a\) at \(t\), given observable state \(x_{a,t}\) (\cref{eq:act})\\
        $\mathvf{a}_t$, $\mathvf{x}_t$, $\mathvf{y}_t$ & Vectors of all agents' states, observable
      agent inputs, and
    agent actions corresponding \(a\) (\cref{eq:io-vectors})\\
    \( \mathrm{act}\)
      & Activation of agent \(a\), transforming the observable system instance
        states \(x_{a,t}\) at \(t\) into an actions \(y_{a,t}\)
        for \(t+1\), altering agent instance states from
        \(\mathvf{a}_t\) to \(\mathvf{a}_{t+1}\) (\cref{eq:act})\\
        \( \mathrm{eval}\)
      & Application of action vector(s) of all agents \(\mathvf{y}_t\)
      at \(t\), causing the world to transite from state \(m_t\) to the state \(m_{t+1}\)~(\cref{eq:eval})\\
    \bottomrule
  \end{tabu}%
  \caption{ARL notation}
  \label{tab:notation}
\end{table}




\subsection{Formal Definition}
\label{sec:formal-definition}

Using the notation introduced here and summarized for reference in
\cref{tab:notation}, we define the concept of \gls{ARL} as model and
connection setup with transition process in the following way.

An setup in \gls{ARL} is comprised of agents $a \in A\cup\Omega$, which include, at
least, one instance of each set $A, \Omega$. Each agent is related to
a set of inputs $X_a$ and a set of outputs $Y_a$.

Further the setup requires a world model $m$ which provides a set of
sensors $X_m$ and controls $Y_m$. 

\begin{figure}
  \centering
  \includegraphics[width=\linewidth]{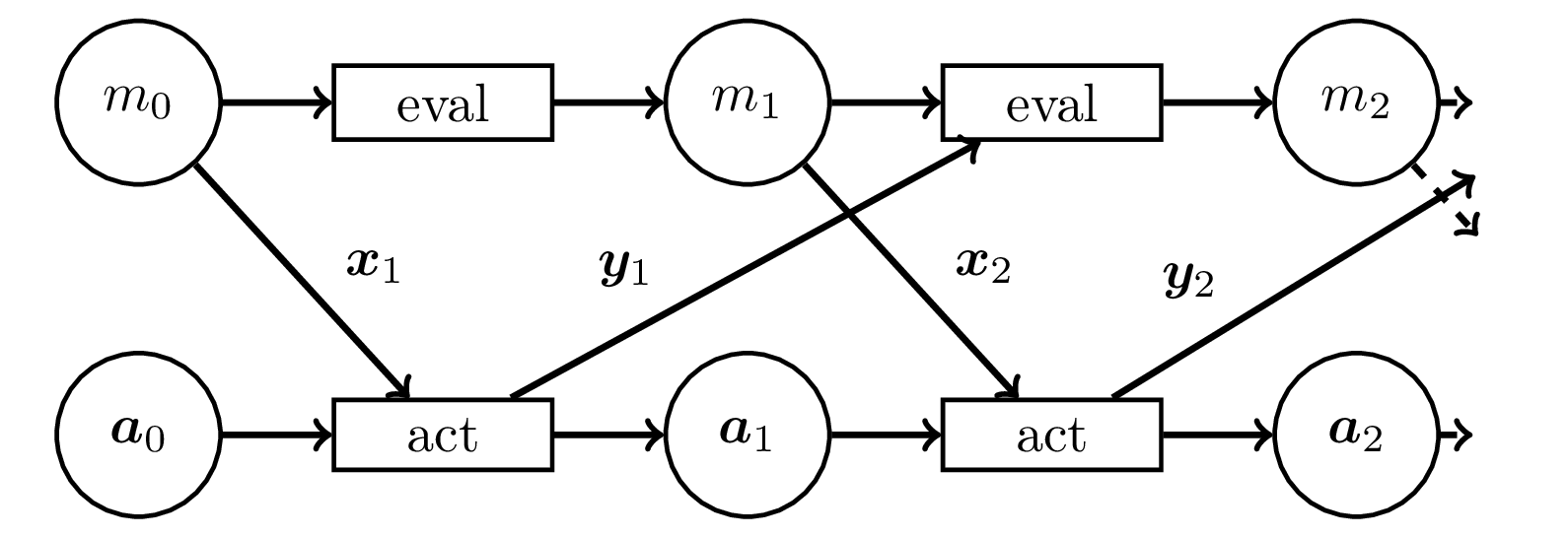}
  \caption{\gls{ARL} sequence of execution}
  \label{fig:arl-sequence}
\end{figure}

The central process of \gls{ARL} is the dynamic system-of-systems view
of a set of agents \(a_0, a_1, \dotsc, a_n\) acting upon a shared
instance of a world model. Activation functions $act_a$ of agents and
application $eval$ of agent agents to a wordl model form a cyclic
sequence of activation and application that transforms the states of
model and agents into a sequence of states as shown in
\cref{fig:arl-sequence}.

An experiment of \gls{ARL} is the execution of this sequence. The
resulting data of an experiment is the sequence of states, inputs and
outputs $(m_t, \mathvf{a}_t, \mathvf{x_t}, \mathvf{y}_t), t\in
\{1,…,n\}$ as well as the initial setup $m_0, \mathvf{a}_0$.

Thus finally we can strive to formalise the idea by collecting all
components in a single scenario:
\begin{definition}[Adversarial Resilience Learning Scenario]
  Any experimental setup comprised of agent instances $a$ of
  $\mA$ of two opposing classes $\mA_A$ and
  $\mA_\Omega$ and a system model
  $\mM$, as well as, for each agent instance
  $a$ a reward function
  $p_a(m,t)$, a mapping of observable states
  $\mathvf{x}_{a,t}$, and action vectors $\mathvf{y}_{a,t}$.
\end{definition}

\Gls{ARL} thus is the application of \gls{RL}, as
introduced in \Cref{sec:machine-learning}, to iteratively
improve the internal decision structure that determines the
behaviour of an agent $act_a$. The output of \gls{ARL} then is,
depending on the exerimenters objectives, an observation of the
performance of the system model $\mM$ or a set of agents trained
towards the defined objectives.

\subsection{Optimization Problem Statement}
\label{sec:problem-statement}

This section describes possible optimization problems that provide the
motivation for \gls{ARL}. 

\begin{figure}
  \centering
  \includegraphics[width=\linewidth]{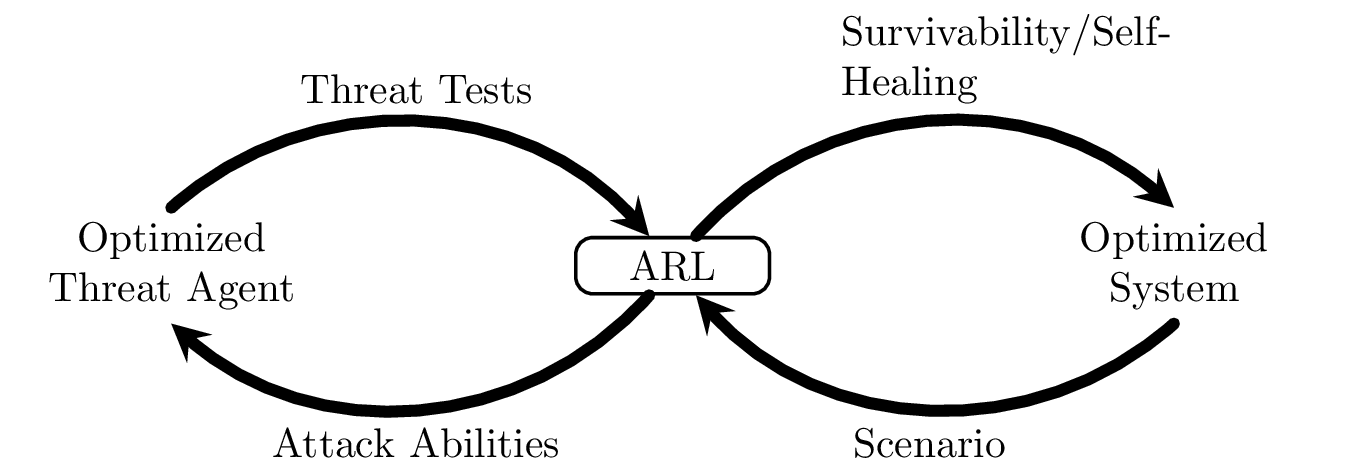}
  \caption{Optimization Objectives}
  \label{fig:training}
\end{figure}

\Gls{ARL} resembles a closed-loop
control situation with (at least) two conflicting controls. Herein are
distinguished two different optimisation objectives that provide
different uses of \gls{ARL}.  The different uses, as depicted in
\cref{fig:training}, improve different elements to achive either an
improved threat tests, or a more resilient system. The primary
distinction is between evolving parameters of ANN in order to optimize
individual agents or step-wise advancing the structure of the system
model. Our concept itself is oblivious to the
algorithms used for optimization.

\subsubsection{System Optimization}
\label{sec:system-optimization}


The primary objective is to find the inherent control asymmetry of a
given control system to finally recommend system designs that favor
the defender over the attacker. In control theory this could be
expressed as a system, where for all possible sequences of actions by
the attacker for a given system model $\mM$ there is at least one
corresponding sequence of actions for the defender and the resulting
performance of the system will never drop below a given failure
threshold. This requirement can be relaxed by defining a finite
measure of failure that may be acceptable, for example during an
initiation phase.

Objectives of defender and attacker in control scenarios are focussed
on system states measured by a model performance function~\cref{eq:model-performance}, as formally given
in~\cref{def:attacker_defender_classes}. In general, we call an agent
\emph{defender} if its objective is to keep the performance at least
above the failure threshold. We denote an agent as \emph{attacker} if
it aims at pushing the performance below a expression for a failure
threshold, as seen in \Cref{fig:resilience}.

We denote the objective of asymmetry — favouring defence of
a system — given a candidate system model instance $m$ and defender
agent $a_\Omega$ that
\begin{equation}
  \label{eq:control_asymmetry}
  p^f < p(m_t) \text{ for all } t > t_0,
  a_A \in A~.
\end{equation}

Which describes that the performance, for all potential attackers $a$ in $A$,
there exists a defender $a_\Omega^*$, with given
initial state $m_{t_0}$, will never fall below a defined failure threshold
$p^f$. To account for a learning period we allow for a finite initialisation
time until $t_{0}$.

Improvement is achieved by evolutionary changes to the system model
$\mM$, improved defensive agent models $\mA$ or training of defensive
agents $a_\Omega$, as discussed in the following section.


\subsubsection{Agent-Training}
\label{sec:experimentation}


Training of threat agents aims at improving attack abilities,
including the identification of previously unknown attack vectors, in
order to provide testing capabilities. Improved threat tests allow to
define test requirements for system designs that improve systems'
resilience against security threats. One objective is to train threat
agents that can be used as benchmarks for future system designs.


An agent's objective is implemented through a \emph{reward function}
that is used within a reinforcement learning process that successively
improves the agent's behaviour towards that objective. 

One particularly surprising success of \glspl{RL}
algorithms has been the identification of solutions unthought-of by
experts, especially if applied to zero-information initial
states.  A two-agent,
conflicting-objectives game only one potential learning structure
usable with \gls{ARL}. But the concept allows potentially for all
combinations of one-or-many zero-information reinforcment agents and
static or even human-controlled competition.

\section{Application to Power Systems}
\label{sec:appl-power-syst}

In this section an example application of \gls{ARL} to adversarial control
in power systems is shown as a feasibility demonstration. We show that
\gls{ARL} provides a novel approach to analyze fundamental control asymmetries
between intelligent attackers and defenders. The final aim of this
application is to design and analyse system configurations that are
inherently favoring the defender in his objective to stabilize system
performance.

Applied on power systems, \eg{} the performance function is expressed
as diversion from a specified range of acceptable state values. The attackers
objective is to force the system to a state where one or more values are
outside allowed ranges, its success is measured by the amount and duration of
the deviation. The ``defender'' has lost the competition if the attacker is
able to divert any of the system's parameters beyond the acceptable range.

Specific objectives for attackers can vary widely as there are many
different parts of a power system that can be affected in order to
disrupt service and reduce system performance. Attackers may aim at
demolition of connected machines or components of the transmission and
control system. Thus, to strive for a more general specification of
objectives, we better consider the objectives of defenders and
specifiy a deviation from these objectives as ``win'' for the
attackers. The objectives for the defender are very well defined in
the power system domain.

\begin{figure}
  \centering
  \includegraphics[width=\linewidth]{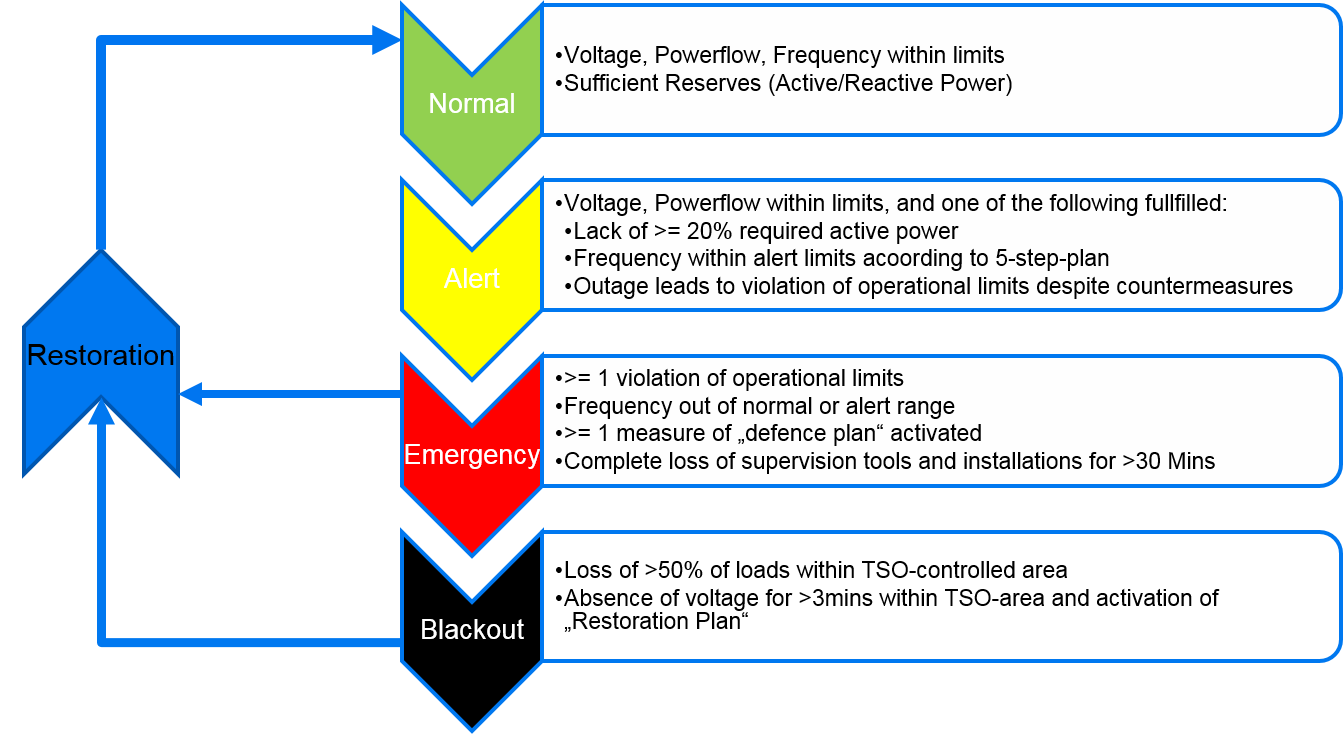}
  \caption{ENTSO-E Operational Phases}
  \label{fig:entso-e-phases}
\end{figure}

Different specific requirements apply for different parts of the power system.
Common parameters to consider are voltage \(V\) in \si{V},\footnote{Voltage
is often given in power unit, \si{pu}, which takes the value 1 for normal
voltage. \Eg{} if \SI{110}{kV} is the nominal voltage in a distribution grid,
\SI{1}{pu} is \SI{110}{kV}, and \(\SI{0.9}{pu} = 0.9 \cdot \SI{110}{kV} =
\SI{99}{kV}\).} frequency \(f\) in \si{Hz}, phasor angle \(\phi\) in \si{rad},
real power \(P\) in \si{W}, and reactive power in \si{VAr}. In general, phase
synchronicity is more important for high-voltage transmission grids, as
asynchronicity leads to harmonics in the power system, with potentially
desastreous large power flows between large segments of the grid.  For the
european transmission grid the operation guidelines, defining conditions for
four phases: normal, alert, emergency, and blackout, shown in \Cref{fig:entso-e-phases}, are defined in \cite{tso_system_operation_2017}.

Similarly operational parameters exist for medium- and
low-voltage-grids, power generation and connected loads.

DIN~EN~50160 specifies parameters for the operation of distribution grids. It
defines that voltage has to stay between \SI{0.9}{pu} and \SI{1.1}{pu}. It is
acceptable, by definition in EN~50160, that voltage drops down to at least
\SI{0.85}{pu} for at most 5\% of a week. Frequency must only deviate from the
nominal 50~Hz by at most 4\% above or 6\% for not more than 0.5\% of the year,
\ie{} less than 2~days overall. Normal operation must deviate no more than
\(\pm 1\)\%.~\cite{en50160}

An attack, in this simulation is deemed successful if any requirement
exceeds its defined limits. 

\begin{figure}
  \centering
  \includegraphics[width=\linewidth]{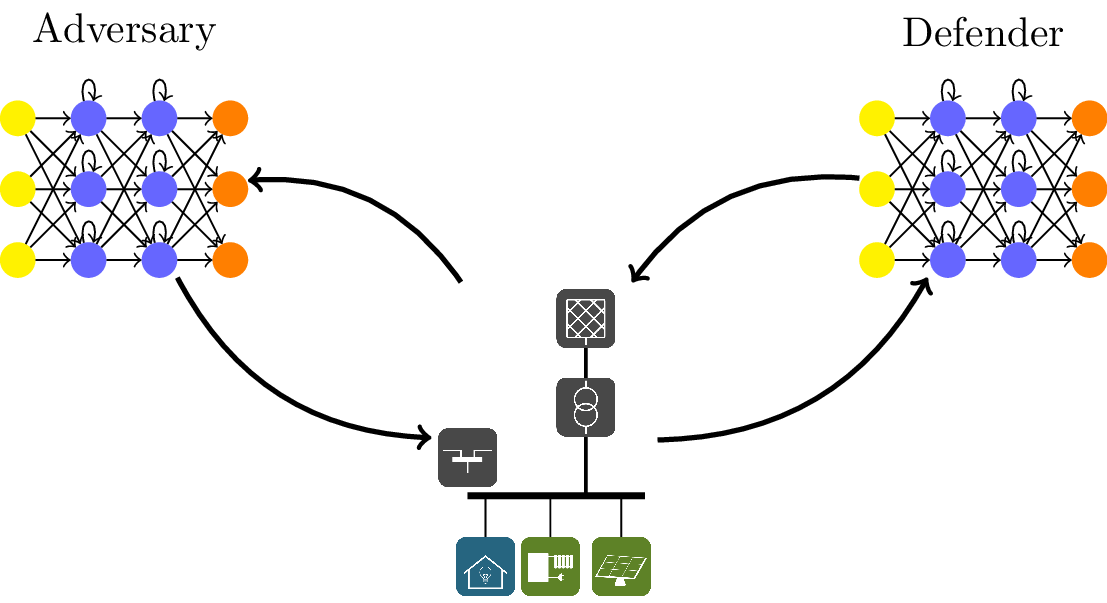}
  \caption{ARL ANN structure}
  \label{fig:ARL-structure}
\end{figure}

\Cref{fig:ARL-structure} shows the refinement of the generic
\gls{ARL}-structure as described in \cref{sec:advers-resil-learn} for
the power grid scenario using \gls{ANN} to implement a single
adversary and a single defender. Both agents interact only through
sensors and actuators that influence different controls in the power grid.

In the remainder of this section we introduce a \gls{PoC} implementation of
ARL using PandaPower~\cite{pandapower.2018} for static grid simulation and the
Keras-RL library \citep{plappert2016kerasrl} for implementation of
reinforcement learning for \glspl{ANN}. First, a brief description of the
control scenario is provided, followed by a discussion of the preliminary
results.

\subsection{Static Control Scenario}
\label{sec:stat-contr-scen}

The objective of this proof-of-concept is to show the general feasibility of
using (multiple) \gls{ANN}-heuristics and train them by reinforcement learning
to modify controls in a static power system simulation towards their
objectives. 

The simulation uses a simple medium voltage power grid, as model from
the static grid simulation pandapower \citep{pandapower.2018}. The
grid contains four generators connected by six transformers to six
loads. For the \gls{PoC}, we chose to only use voltage as
state-indicator and input to the reward of the attacker. The initial
configuration of the grid comprises a stable healthy state of the grid
that would be held up constantly if no control actions would be
initiated.


Actuators in this scenario are tap changer, reactive power control,
loads and generation levels as represented by the commonly deployed
and future automated controls in power systems.

The reward function for the attacker is shown at the bottom right in
\Cref{fig:attacker_reward}. Initial trials pointed towards the inverse of a Poisson
Density Function centered on the nominal voltage unit. The reward function
thus resembles the objective for an attacker, providing only positive rewards
if the mean voltage deviates more than 5\% from the nominal voltage.  The
single agent in this demonstration had been assigned direct control of every
transformer, generator and load in this scenario.

In terms of optimization from \Cref{sec:problem-statement}, the
scenario instantiates $m$ from $\mM = \text{ Simple Example}$, with a
single agent $a\in \mA_A$, with a parametrized normal distribution
\begin{equation}
  \label{eq:reward_function}
  p_a(m,t) =
  {-1}^{[a\in\mA_\Omega]} e^{-\frac{(x-\mu)^2}{2\sigma^2}} - c,
\end{equation}
where $c, \mu$ and $\sigma$ parametrize the reward curve,
${-1}^{[a\notin\mA_A]}$ negates the reward if $a$ is an
attacker\footnote{Where $[]$ denote Iverson brackets \cite{Iverson1962}}, and
$x = \mathop{mean}{\mathvf{x}_{a,t}}$ is the average of all inputs.

\subsection{Demonstrator}
\label{sec:demonstrator}

In order to show the genearal feasibility of the concept, we
implemented a demonstrator for reinforcement learning in power control
scenarios. The current implementation uses static simulation in
PandaPower~\citep{pandapower.2018}. It supports free
configurability of controlled sensors and actuators of multiple
agents, selection of ANN-algorithms and -parameters, as well as
different logging and output formats. 

\begin{figure}
  \centering
  \includegraphics[width=\linewidth]{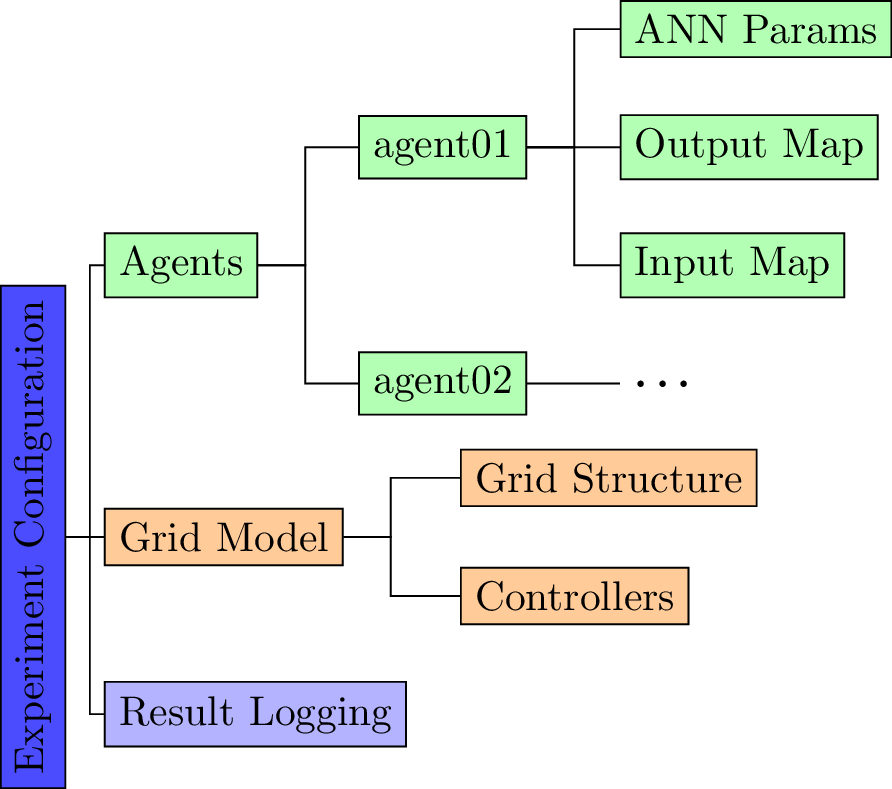}  
  \caption{ARL Demonstrator configuration structure}
  \label{fig:arl-config}
\end{figure}

Each experiment is specified within a single configuration file, in
order to support documentation and reproducibility of experiments. A
experimental configuration (\Cref{fig:arl-config}) defines three
major simulation components: a grid model, one or more agents, and a
collection of result logs that collect results. At the time of writing,
the whole demonstrator is refactored to use the mosaik co-simulation
framework.~\cite{Schutte2013,Lehnhoff2015}

The interconnection between agents and grid simulation, \ie{} the
inputs (sensors) and outputs (actuators), $\mathvf{x}_a$ and
$\mathvf{y}_a$ respectively, are separately defined for each agent.

The execution of the simulation is round based. The rounds are
advanced in steps according to a defined evaluation order of
agents. Agents are sequentially executed, a defined number of steps
each. The grid state is evaluated between each consecutive pair of
agent evaluation steps. After each step, the internal weights of an
agent are modified by the learning algorithm selected in the
configuration of an agent.

Current result monitors output the grid states at every node of the
grid into a grid-state-log. For the agents, a log consisting of
inputs, outputs and evaluated reward for the output is output to a
configurable file in CSV format. The results are graphically evaluated
as is discussed in \Cref{sec:results} below.

\subsection{Results}
\label{sec:results}

\begin{figure*}
  \centering
  \begin{subfigure}[t]{.5\linewidth}
    \includegraphics[width=\linewidth]{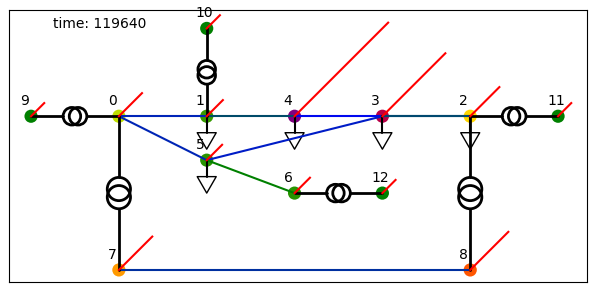}
    \caption{Grid Simulation}
    \label{fig:grid_sim}
  \end{subfigure}
  \begin{subfigure}[t]{.4\linewidth}
    \includegraphics[width=\linewidth]{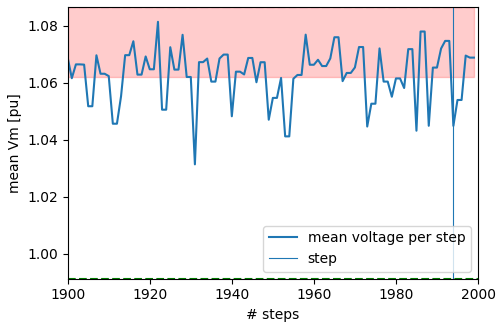}
    \caption{Grid performance (mean voltage) over time/evaluation steps}
    \label{fig:grid_voltage}
  \end{subfigure}
  \caption{Proof-of-concept \gls{ARL} grid results}
  \label{fig:arl-grid-results}  
\end{figure*}

\begin{figure*}
  \begin{subfigure}[T]{.3\linewidth}
    \includegraphics[width=\linewidth]{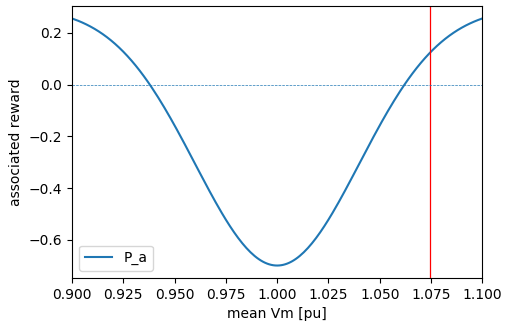}
    \caption{Attacker reward function, based on mean voltage}
    \label{fig:attacker_reward}
  \end{subfigure}
  \begin{subfigure}[T]{.3\linewidth}
    \includegraphics[width=\linewidth]{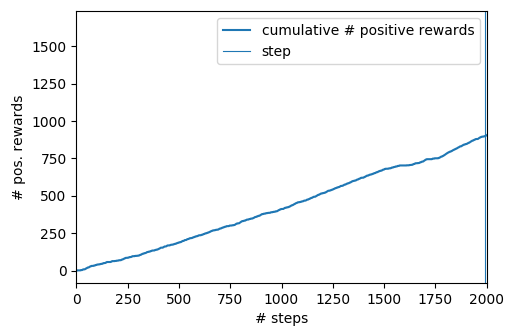}
    \caption{Cumulative number of positive rewards over training steps}
    \label{fig:attacker_cumulative}
  \end{subfigure}
  \begin{subfigure}[T]{.3\linewidth}
    \includegraphics[width=\linewidth]{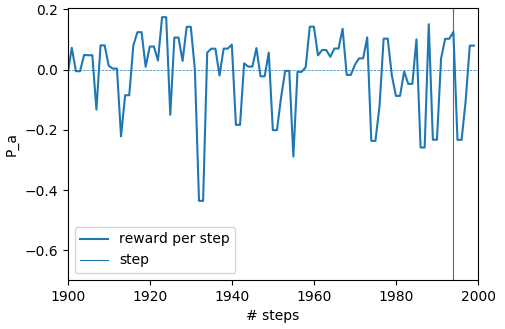}
    \caption{Attacker performance $p_a(m_t)$ over time (inverse mean voltage)}
    \label{fig:attacker_performance}
  \end{subfigure}
  \\
  \begin{subfigure}[t]{.3\linewidth}
    \includegraphics[width=\linewidth]{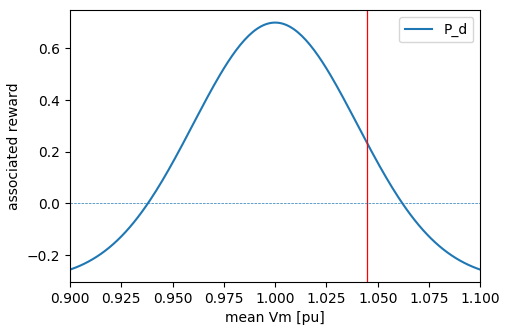}
    \caption{Defender reward function, based on mean voltage}
    \label{fig:defender_reward}
  \end{subfigure}
  \begin{subfigure}[t]{.3\linewidth}
    \includegraphics[width=\linewidth]{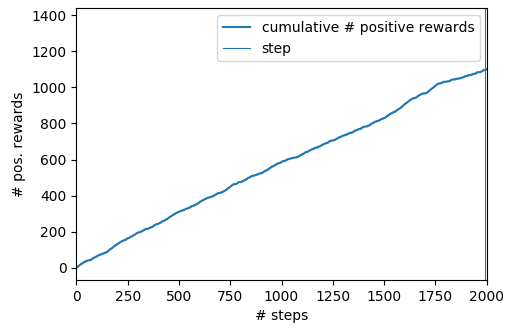}
    \caption{Cumulative number of positive rewards over training steps}
    \label{fig:defender_cumulative}
  \end{subfigure}
  \begin{subfigure}[t]{.3\linewidth}
    \includegraphics[width=\linewidth]{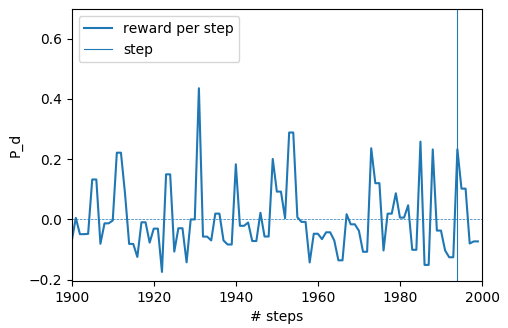}
    \caption{Defender performance $p_a(t)$ over time (mean voltage)}
    \label{fig:defender_performance}
  \end{subfigure}
  \caption{Proof-of-concept \gls{ARL} agent results}
  \label{fig:arl-agent-results}
\end{figure*}

To show the usability of our demonstrator we pitched two very
simplistic agents with inverse reward functions
(\Cref{fig:attacker_reward} and
\Cref{fig:defender_reward}) against each other, using the
example grid shown in \Cref{fig:grid_sim} as an arena. Both
agents were assigned all voltage sensors as input and we divided the
actuators among their outputs. The attacker was assigned control of
all tab changers, representing a scenario where a vulnerability in one
type of controller was exploited. The defender would be granted access
to all generators and loads in this scenario.


\Cref{fig:arl-grid-results} shows a late state of the
simulation. Seemingly the attacker gained the upper hand and has been
able to increase voltage levels beyond 1.05~pu. The grid
representation in \Cref{fig:grid_sim} shows that especially 
two central measure points (numbered 4 and 3) are struck with very
high voltage levels, represented by the length of the bars rooted at
the nodes, most likely sufficent for the connected loads to
shut down or be damaged. The mean voltage level of the system, depicted for
steps 1900 until 2000, in
\Cref{fig:grid_voltage} shows that even the lower voltages of
other nodes, \eg{} 11 and 9 are not sufficient to lower the mean
voltage to acceptable levels. Thus, in this example the attacker has
been able to destabilize the grid, despite the efforts of the attacker.

Evaluating the two agents in
\Cref{fig:arl-agent-results} provides no immediately conclussive cause for
the loss of the defender. The cumulative number of positive rewards in
\Cref{fig:attacker_cumulative} for the attacker and
\Cref{fig:defender_cumulative}, show only small
differences. These asymmetries might be explained by the order of
execution, where the defender always acts in response to the
attacker. The current reward for the depicted step in the simulation, depicted in
\Cref{fig:attacker_reward} and \Cref{fig:defender_reward},
shows that the defender is evaluating a different mean voltage than
the attacker. Each reward is calculated after the actions of an agent,
thus this graphs show the results of two actions that both improved
the performance towards their own objectives.

In this simulation run, that contained a small random input to the
learning algorithm, the positive learning curve for both agents, in
\Cref{fig:attacker_cumulative} and
\Cref{fig:defender_cumulative}, is increasing right from the
start of the simulation. In preliminary tests with a lone attacker,
the learning process first went through a lengthy phase where only
little positive rewards where achieved. Although we have no
conclussive answer for the reasons of this diverging learning
behaviour, it shows at least, that the interactions between
adversaries have an effect on their behaviour.

\section{Discussion and Future Work}
\label{sec:disc-future-work}

\glsresetall


\todolaf{Non-commutative application of actions}

This work introduced \gls{ARL}, a novel approach to analyse
competetive situations in highly-complex systems using reinforcement
learning with artificial neural networks as reproducable,
self-improving agents. The concept can be seen as an extension of
\gls{GAN}, which distinguishes itself by
including interaction to a complex system (simulation) in
between the competing agents. This work is motivated by the need to
find better methods to automatically evaluate system behaviour under
threat of maliciously acting, intelligent threat agents. The main idea
is, that groups of agents, modelled by \gls{ANN}, struggle to enforce
their objectives against agents with conflicting objectives. 


Pitching two---or more---\glspl{ANN} with conflicting reward functions against
each other may allow to define more realistic tests for adversarial or
competitive situations. Using reinforcement learning harbours the promise of
finding novel strategies for both attack and defense, which both can be used
to strengthen the resilience of systems during the design and testing phase of
a power system or individual components. \gls{ARL}-based analysis should
contribute to building grid structures that are more resilient to attacks and
train both artificial and human operators in better handling of security
incidents.

Generally, the concept may allow to estimate threat-related indices, for
example the maximum amount of control that an adversary may be allowed to gain
over a system, which leads to improved and more effective recommendations for
security directives and risk mitigations.


Furthermore, we believe that \gls{ARL} provides a valuable addition to
the increased complexity for adversarial learning. Agents in \gls{ARL} have
not only to approximate the behaviour of a highly-complex systsem, but
also have to learn and adopt to changing behaviour of this system due
to the actions inflicted by the competing agents. In this way, not
only the amount of control over a system provides a source of
asymmetry, but also power of the \gls{ANN}.

The concept of \gls{ARL} and its ongoing implementation in the
\gls{ARL}-Demonstrator only marks the starting point for in-depth
research on structural asymmetries of complex systems and protection
against learning threat agents. The demonstrator provides the
abilities to further research in a number of interesting directions.

Foremost is the analysis of structural resilience of complex
systems, especially finding minimum control sets of critical
components that provide the most defensive capabilities, or estimates
of the structural strength of a system. The integration into our
co-simulation framework mosaik opens up the possibility of extending
the single system into a whole composition into an interdependent
system-of-systems. Introduction of multiple domains would, to give one
example, allow to
analyse the effects of \gls{ICT}-components onto the performance of
power systems. 

Deeper extensions of the demonstrator itself will involve capabilities
of the defender to affect structural changes to the system. This would
allow to use \gls{RL} to identify novel and more resilient
structures. The dual ability for threat agents would be the extension
of control, \ie{} simulation of further compromise from within a
system. Both activities require the introduction of a measure of cost
to the demonstrator. 


This demonstrator further allows to analyse simulated systems from the point
of view of threat agents, by pitching the agent against novel security
measures, for example simulation of distributed coordinated attacks. Combining
this view with multi-domain scenarios, would enable analysis of sophisticated,
multi-level attack techniques that involve, for example information hiding or
emission of misleading information by attacker or defender. That means finding
novel ways of attack using a combination of illeagal and legal operations and
interdependencies between different systems. Consequentially all these
approaches would lead to the development of improved designs and testing
methods for highly complex systems.

We can only assume that this finally lead to more resilient designs
and defensive adaptable strategies — and, in the end, to improvements
for the security of supply, but at this stage of the work, the first
results are very satisfying.


\bibliographystyle{ACM-Reference-Format}
\bibliography{arl-references}

\end{document}